\documentclass[journal]{IEEEtran}
\usepackage{amsmath,amssymb}
\usepackage{booktabs}
\usepackage{graphicx}
\usepackage{multirow}
\usepackage{array}
\usepackage{tabularx}
\usepackage{threeparttable}
\usepackage{enumitem}
\usepackage{microtype}
\usepackage{xcolor}
\usepackage{cite}
\usepackage{url}
\usepackage{balance}
\usepackage{stfloats}
\newcolumntype{Y}{>{\raggedright\arraybackslash}X}
\newcolumntype{L}[1]{>{\raggedright\arraybackslash}p{#1}}

\newcommand{\Mhat}{\widehat{M}}
\newcommand{\Mstar}{M^{*}}
\newcommand{\FA}{\mathrm{FA}}
\newcommand{\RA}{\mathrm{RA}}

\newcommand{\MIAadv}{\mathrm{MIA}_{\mathrm{adv}}}
\newcommand{\AD}{\mathrm{AD}}
\newcommand{\JS}{\mathrm{JS}}
\newcommand{\tauB}{\tau_{b}}

\newcommand{\Fw}{VGUA}
\newcommand{\FwLong}{validity-gated unlearning audit}

\title{Validity-Gated Evaluation of Multimodal Machine Unlearning}
\author{Abdullah Ahmad Khan, Hamid Laga, and Ferdous Sohel, \IEEEmembership{Senior Member, IEEE}
\thanks{The authors are with the School of Information Technology, Murdoch University, Perth, WA 6150, Australia. Corresponding author: Ferdous Sohel (e-mail: f.sohel@murdoch.edu.au).}}

\begin{document}
\maketitle

\begin{abstract}
Machine unlearning aims to remove the influence of training data without retraining a model from scratch. Evaluation in vision--language models relies on forget accuracy (FA), retain accuracy (RA), membership inference, and similarity to a retain-only reference, but evaluation pipelines are rarely tested. We propose a \FwLong{} (\Fw{}) with three checks: the forget set must be memorized without duplicate leakage into the retain set, the scorer must measure correctness, and the training code must implement the intended procedure. Applying these checks reveals that random MMBench splits leak duplicate questions, a lenient scorer hides memorization, and two coding errors remove the signal while training completes. After correcting these problems, we evaluate 234 unlearned models using six methods, three learning rates, and three backbones. Each model is compared with a retain-only reference trained using low-rank adaptation (LoRA). The results depend on update strength. Among models with RA at or above 0.70, lower FA generally indicates closer agreement with the retrained FA--RA operating point. Aggressive updates overshoot this point, collapse RA, and reverse method rankings. Relative to the original model, 78\% of healthy models move closer to the retrained scores, whereas 94\% move farther from the retrained model in activation space. Membership-inference advantage does not reliably track retraining alignment. Under option-shuffled training, unlearned models answer 25\% of reordered forgotten ScienceQA questions, compared with 6\% for the retain-only reference. These results show that multimodal unlearning conclusions depend on pipeline validity, update strength, and the comparison criterion.
\end{abstract}

\begin{IEEEkeywords}
Evaluation reliability, machine unlearning, memorization, multimodal learning, pipeline validity, vision--language models.
\end{IEEEkeywords}
\section{Introduction}
\label{sec:intro}

Trained models are increasingly expected to forget selected training data on request. These requests can arise from privacy laws, copyright withdrawal, data-governance requirements, and safety updates. Retraining a model from scratch for every request is usually too expensive. Machine unlearning provides an alternative by updating a trained model to remove the influence of selected examples~\cite{cao2015machine,bourtoule2021machine,sekhari2021remember,nguyen2022survey}. However, deciding whether an unlearning method has succeeded remains an open problem. Recent surveys identify evaluation, verification, and standardization as major unresolved issues~\cite{shaik2025tnnls,li2025tnnls}. Similar concerns also arise in federated unlearning~\cite{romandini2025federated}. Evaluation has therefore become a weak part of the unlearning pipeline.

The problem is more difficult for vision--language models (VLMs). A VLM can store the same information in its visual features, textual features, cross-modal projections, low-rank adaptation (LoRA) parameters~\cite{hu2022lora}, hidden activations, and output distribution~\cite{liu2023llava}. Removing information from one part of the model does not guarantee its removal from the others. For example, a model may stop producing the target answer while an attacker can still identify that the forgotten examples were used during training~\cite{hayes2025inexact}. Similarly, its outputs may resemble those of a retrained model while its internal activations remain different~\cite{kim2026truly}. Evaluating forgetting can therefore be as difficult as performing it.

Recent multimodal unlearning research can be divided into two groups. The first group develops benchmarks and scoring methods. For instance, MLLMU-Bench studies the removal of fictitious person profiles~\cite{liu2025mllmu}. Single-image unlearning studies the removal of visual concepts~\cite{li2024siu}. MMUnlearner and SafeEraser examine the balance between forgetting and preserving useful knowledge~\cite{huo2025mmunlearner,chen2025safeeraser}. CLEAR studies information removal across both text and images~\cite{dontsov2025clear}. These benchmarks expand the range of information that can be forgotten, but they usually assume that the evaluation pipeline is reliable.

The second group studies whether apparent forgetting is genuine. For example, SUA shows that targeted visual changes can recover knowledge that was considered forgotten~\cite{zhang2025sua}. ReMem shows that an unlearning benchmark is difficult to interpret when the selected information was not learned well before unlearning~\cite{kwon2026remember}. GEMMU shows that the best-performing method can change when the learning rate changes~\cite{chew2026gemmu}. These studies demonstrate that a good unlearning score can be misleading. However, they do not examine whether commonly used evaluation criteria agree with the retrained model that unlearning aims to approximate. A recent survey also treats deletion strength, utility, robustness, and reversibility as different evaluation dimensions~\cite{sarwar2026survey}.

Three problems remain. First, an unlearning score has little meaning if the forget set was not memorized before unlearning. The score may also be misleading if duplicate questions appear in both the forget and retain sets. Second, an incorrect scorer or a coding error can hide the signal being measured. Third, it is unclear whether commonly reported criteria agree with the retrained model. This agreement may also change with the strength of the update.

In this paper, we use \emph{evaluation reliability} to describe how well an evaluation score can be trusted to rank unlearned models. We use \emph{operating regime} to describe the strength of the update, which is controlled here by the learning rate. We call an unlearned model \emph{healthy} if it still performs well on the retained data, that is, its retain accuracy stays at or above 0.70. We call it \emph{collapsed} if unlearning also destroys this performance and retain accuracy falls below 0.70 (Section~\ref{sec:dose}). This paper asks:

\begin{quote}
\emph{When an unlearning study ranks one unlearned model above another, how much does this ranking depend on the validity of the evaluation pipeline, the operating regime, and the selected evaluation criterion?}
\end{quote}

\begin{figure*}[t]
\centering
\includegraphics[width=\textwidth]{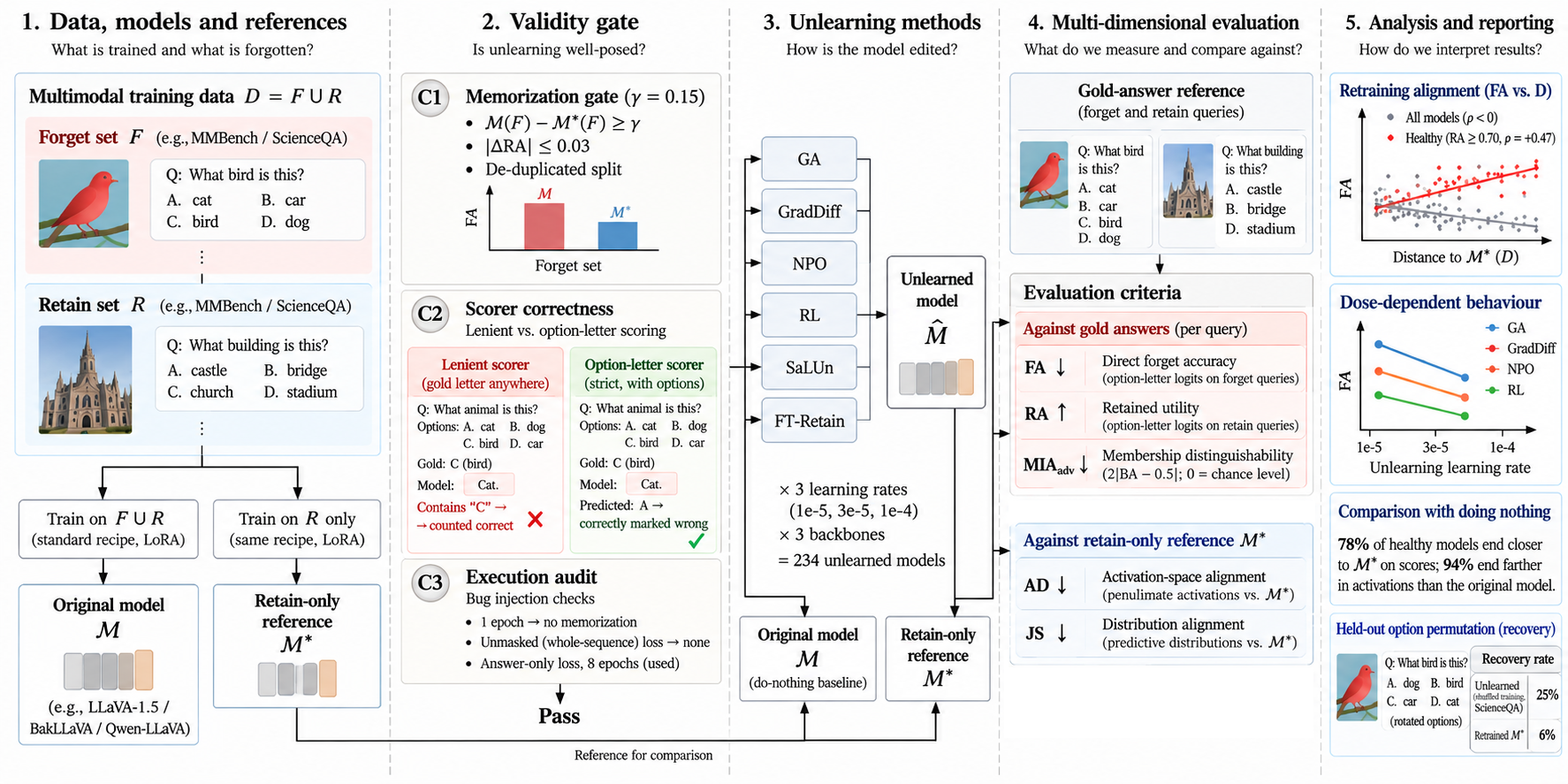}
\caption{Overview of the \FwLong{} (\Fw). 
(1)~The original model $M$ is trained on $\mathcal{D}=F\cup R$ using low-rank adaptation (LoRA). The retain-only reference $\Mstar$ is trained on $R$ using the same settings. 
(2)~Three conditions are checked before the unlearning results are interpreted. Conditioning (C1) checks memorization, retain stability, and duplicate leakage. Measurement (C2) checks whether the scorer correctly identifies the selected answer. Execution (C3) checks whether coding errors change the intended training procedure. 
(3)~Six methods, three learning rates, and three backbones produce 234 unlearned models $\widehat{M}$. 
(4)~Each model is evaluated using forget accuracy (FA), retain accuracy (RA), membership-inference advantage ($\MIAadv$), activation distance (AD), and Jensen--Shannon divergence (JS). The original model $M$ provides the do-nothing baseline. 
(5)~The analysis tests whether these criteria track the distance $D$ to the retrained FA--RA operating point. It also examines healthy and collapsed models, changes in method rankings, activation-space alignment, and the recovery of forgotten answers after option reordering.}
\label{fig:framework}
\end{figure*}

\noindent To study this question, we propose a \FwLong{} (\Fw), rather than another unlearning algorithm (Fig.~\ref{fig:framework}). \Fw{} begins with three validity checks. The conditioning check requires the forget set to be memorized before unlearning. It also checks that duplicate questions are not shared between the forget and retain sets. The measurement check determines whether the scorer correctly identifies the model's answers. The execution check determines whether the training code performs the intended procedure. We test execution by introducing realistic coding errors and measuring their effects.

Only settings that pass all three checks are included in the main study. We evaluate six unlearning methods at three learning rates on MMBench~\cite{liu2024mmbench} and ScienceQA~\cite{lu2022scienceqa}. We use three backbones. LLaVA-1.5-7B~\cite{liu2024visual} and BakLLaVA~\cite{bakllava2023} are evaluated with three seeds each. A Qwen-based LLaVA model~\cite{li2024llavanext} is evaluated with one seed. This produces 234 unlearned models. We also evaluate 36 additional models after training LLaVA with shuffled answer-option orders.

For each unlearned model, we measure forget accuracy (FA) and retain accuracy (RA). We compare these scores with those of a matched reference model trained only on the retained data, which serves as the retraining reference. We compute the distance $D$ between the FA--RA scores of each unlearned model and those of the reference. We also compare each unlearned model against the original model, which serves as a do-nothing baseline, as it receives no unlearning update. All reported results come from settings that pass the three validity checks.

The contributions are of two kinds: methodological contributions and empirical findings. The first two contributions describe the proposed evaluation method.

\begin{enumerate}[leftmargin=1.6em,label=\arabic*),itemsep=2pt,topsep=3pt]

\item \textbf{Validity conditions.} We define three testable conditions: a memorization, retain-stability, and duplicate-leakage check; a scorer check; and an execution check.

\item \textbf{A retraining-aligned evaluation design.}
We assess each evaluation criterion by testing whether it tracks the distance $D$ to the retrained FA--RA operating point. This operating point is a standard target in classification unlearning~\cite{jia2023sparsity,fan2024salun,triantafillou2024progress}. The design includes a do-nothing baseline and separates healthy models from collapsed models. It also includes held-out and option-shuffled evaluations.

\end{enumerate}

\noindent The remaining three contributions report the main empirical findings.

\begin{enumerate}[leftmargin=1.6em,label=\arabic*),itemsep=2pt,topsep=3pt]
\setcounter{enumi}{2}

\item \textbf{The validity conditions fail in realistic pipelines.}
Under our training recipe, MLLMU-Bench fails the memorization check. Its memorization gap is 0.13, below the required threshold of 0.15. MMBench contains 4{,}377 rows but only 1{,}175 unique questions. A random split therefore places about half of the forget questions in the retain set. A lenient scorer and token-level F1 also hide most of the memorization signal. Two realistic coding errors produce a similar problem. These errors are training for only one epoch and using an unmasked loss. In both cases, the training run finishes normally even though forget-specific memorization is lost.

\item \textbf{The meaning of FA depends on the operating regime.}
Among healthy models, lower FA usually means that the model is closer to the retrained FA--RA operating point. However, very strong updates can reduce FA beyond the retrained target and cause RA to collapse. In this collapsed regime, FA ranks the unlearning methods in the opposite order. Compared with the do-nothing baseline, 132 of the 170 healthy models move closer to the retrained FA--RA scores. However, 159 of these models move farther from the retrained model in activation space.

\item \textbf{Membership inference and recoverability provide different evidence.}
Membership-inference advantage ($\MIAadv$) does not track the distance $D$ to the retraining reference. Across unlearned models, the Spearman correlation between $\MIAadv$ and $D$ is $\rho=-0.14$ when the five model--benchmark settings are combined by a random-effects meta-analysis, and its 95\% confidence interval $[-0.33, 0.06]$ includes zero. A lower $\MIAadv$ therefore does not indicate a model closer to retraining. The option-shuffled experiment also shows that apparently forgotten answers can remain recoverable. On ScienceQA, unlearned models answer 25\% of forgotten questions after the answer options are reordered, whereas the retrained reference answers only 6\%.

\end{enumerate}

\noindent \Fw{} is an evaluation method, not a test that certifies complete erasure. A low FA does not prove that the selected information has been erased. A low membership-inference advantage only provides evidence against the particular attack that was tested. The retrained model is a useful reference, but it is not the only possible definition of successful unlearning. The quantitative results are limited to the LLaVA-family models and experimental settings studied in this paper (Section~\ref{sec:limitations}).

The remainder of this paper is organized as follows. Section~\ref{sec:related} reviews related work. Section~\ref{sec:ontology} defines the evaluation criteria, the retraining target, and the validity conditions. Section~\ref{sec:protocol} presents the validity checks and the main experiments. Section~\ref{sec:discussion} discusses the findings, presents a reporting protocol, and explains the limitations. Section~\ref{sec:conclusion} concludes the paper.

\section{Related Work}
\label{sec:related}

\subsection{Machine Unlearning and Its Evaluation}

Machine unlearning (MU) seeks to remove the influence of designated training data without the expense of retraining a model from scratch
~\cite{cao2015machine,bourtoule2021machine,sekhari2021remember}. Neural unlearning methods take different routes to this goal, including gradient-based forgetting, fine-tuning on retained data, saliency-restricted updates, preference-based objectives, and teacher-student procedures
~\cite{golatkar2020eternal,wu2020deltagrad,fan2024salun,zhang2024npo,kurmanji2024towards}, and second-order methods use Hessian information to approximate retraining at lower cost~\cite{KHAN2026135115}. Target-level unlearning extends removal beyond instance deletion by identifying the parameters that influence a requested target while protecting those needed by the remaining targets~\cite{xu2025essential}. The mechanisms differ, but they leave the same question afterward: did the model forget the requested information while retaining the knowledge that should remain?

Recent work makes this evaluation requirement explicit. Shaik \emph{et al.} organize MU around data- and model-centric mechanisms and identify evaluation, attacks, standardization, interpretability, and resource constraints as unresolved issues~\cite{shaik2025tnnls}. Li \emph{et al.} distinguish among effectiveness, efficiency, and utility metrics~\cite{li2025tnnls}. They note that many measures of effectiveness compare the unlearned model with a retrained model, whereas accuracy alone provides only coarse evidence. In image classification this comparison is standard practice: Jia \emph{et al.} and SalUn rank methods by the average gap to the retrained model in forget, retain, test, and membership accuracy~\cite{jia2023sparsity,fan2024salun}, and the first NeurIPS unlearning competition scored submissions by how closely their output distributions matched those of retrained models~\cite{triantafillou2024progress}. We adopt the same principle as our target. Our question is different: whether the criteria that multimodal studies usually report, forget accuracy read as lower-is-better and membership inference, agree with that target, and in which regime they fail. Romandini \emph{et al.} report a similar problem in federated unlearning, where forgetting is verified in inconsistent ways~\cite{romandini2025federated}. More recently, Xue \emph{et al.} organize unlearning verification into behavioral and parametric approaches and highlight the assumptions and limitations associated with different forms of verification evidence~\cite{xue2026reliable}, and Hayes \emph{et al.} show that membership-inference evaluations of inexact unlearning can give a false sense of privacy when the attack is not adapted to each example~\cite{hayes2025inexact}.

\subsection{Multimodal Representations and Multimodal Unlearning}

Multimodal models make verification more difficult because the information being removed may not be confined to the final prediction. Work on multimodal metric learning, cross-modal distillation, and parameter-efficient VLM adaptation shows that information is distributed across cross-modal representations and adapter components~\cite{wang2024dhmml,tan2025vln,li2025csp}. For unlearning, this means that final answer accuracy captures only one part of the change: outputs, hidden activations, predictive distributions, and low-dimensional adaptation parameters may respond differently to the same deletion update.

Multimodal unlearning is not evaluated under a single notion of forgetting. The target may be an entity, as in MLLMU-Bench~\cite{liu2025mllmu}, or a visual concept, as in single-image unlearning~\cite{li2024siu}. Other evaluations ask how much useful information survives the deletion; MMUnlearner and SafeEraser explicitly study this forgetting--preservation trade-off~\cite{huo2025mmunlearner,chen2025safeeraser}. CLEAR adds another dimension by testing removal across textual and visual modalities~\cite{dontsov2025clear}. Even these measurements may miss information that remains accessible in the model: it can be recovered under visual perturbations~\cite{zhang2025sua}, through relearning and prompt attacks~\cite{zheng2026offside}, or from intermediate reasoning~\cite{li2026reasoning}. There is also a problem before unlearning begins: ReMem shows that the resulting scores are difficult to interpret when the target information was not reliably learned in the first place~\cite{kwon2026remember}. Our memorization gate turns that observation into a pass/fail test applied before any unlearning result is reported.

\subsection{Evaluation Reliability and Closest Work}

Benchmarks such as TOFU, MUSE, and OpenUnlearning already evaluate generative models using several behavioral criteria
~\cite{maini2024tofu,shi2024muse,dorna2025openunlearning}. At the same time, recovery studies show that information which appears to have been forgotten can return after alternate prompting, fine-tuning, quantization, or targeted interventions
~\cite{lynch2024eight,rezaei2025restor,zhang2025catastrophic,sun2025obfuscation}. Recent work has also questioned whether conventional output-level evaluation is sufficient. Kim \emph{et al.} show in image classification that favorable output-level forgetting can coexist with unlearned models that remain internally closer to the original model than to the retrained one~\cite{kim2026truly}, and Seo \emph{et al.} compare the latent feature structure of original and retrained models to expose weaknesses that conventional criteria miss~\cite{seo2025dimensional}. Gradient-ascent-style updates are also known to collapse the model when they are run too hard~\cite{zhang2024npo}, so the strength of the update is part of the result. Feng \emph{et al.} show that LLM unlearning evaluations can be inconclusive because the evaluation itself injects information or depends on the task~\cite{feng2025inconclusive}, and Deep Unlearn finds that method comparisons in image classification depend on hyperparameter selection~\cite{cadet2024deepunlearn}. Having several measurements, however, does not tell us whether they support the same conclusion.

The individual ingredients of our evaluation are not new; our contribution is to test the pipeline that produces the measurements and then ask, on the same multimodal checkpoints, whether the criteria agree with one counterfactual across operating regimes. FIUBench verifies in a learning stage that fictitious identities were acquired before unlearning and compares unlearned models with a model fine-tuned on the retain set only~\cite{ma2025fiubench}; DUSK states directly that an unlearned model should be indistinguishable from one retrained on the retain set~\cite{jeung2026dusk}; ReMem shows that under-memorized targets make LVLM unlearning benchmarks unreliable~\cite{kwon2026remember}; OpenUnlearning meta-evaluates the faithfulness and robustness of text-only unlearning metrics~\cite{dorna2025openunlearning}; and Thaker \emph{et al.} argue that unlearning benchmarks give an overly optimistic view of progress~\cite{thaker2024position}. ULTRA combines behavioral, membership, representation, and retain-consistency requirements into a post-unlearning certification with worst-case aggregation~\cite{arslan2026ultra}; \Fw{} instead treats conditioning, measurement, and execution as preconditions for interpreting any score and keeps the metric vector rather than issuing an erasure certificate. We do not claim that pre-unlearning memorization checks or retain-only references are new. Our contribution is to combine them with scorer and execution audits and test metric rankings across operating regimes. We turn the first into a fixed pass/fail gate defined against the retain-only model, add the scorer and the training code as further validity conditions, and use the reference to test whether commonly reported criteria order unlearned models by their proximity to it.

Closest to our setting, GEMMU~\cite{chew2026gemmu} identifies three pitfalls in multimodal unlearning benchmarks: evaluating on the examples used for unlearning instead of held-out ones, targeting fine-tuned rather than pre-trained knowledge, and comparing methods under one shared configuration; under shared learning rates the winning method flips. Our study differs from GEMMU in its reference. GEMMU targets pre-trained knowledge, for which no retrained model exists, and compares methods by forget and retain drops. We target fine-tuned knowledge precisely so that a matched retain-only reference can be trained with the same adaptation recipe, and we ask whether each criterion moves with the distance to that model; the memorization gate, scorer audit, and bug injection test the pipeline that produces those numbers. The two designs are complementary: GEMMU shows that rankings depend on shared hyperparameters, and we show in which direction FA misleads when compared with retraining. The next section defines these measurements.

\section{Problem Formulation and Evaluation Framework}
\label{sec:ontology}

Let $\mathcal{D}=F\cup R$ denote the training data, where $F$ contains the data to be forgotten and $R$ contains the data to be retained. A base model $M$ is trained or fine-tuned on both sets, and an unlearning algorithm produces $\Mhat=\mathcal{U}(M,F,R)$. We compare this unlearned model with a retain-only reference $\Mstar$, trained on $R$ alone. $\Mstar$ is not a gold-standard oracle. The retain data may contain information correlated with $F$, the reference depends on one adaptation recipe, and other valid retain-only solutions may exist.

We define the criteria (Section~\ref{sec:criteria}) and the target used to assess them (Section~\ref{sec:target}). We then state the validity conditions a score must meet before it is interpreted (Section~\ref{sec:validity}).

\subsection{Automatic Evaluation Criteria}
\label{sec:criteria}

\paragraph{Forget Accuracy (FA)}
For a forget evaluation set $F_{\mathrm{eval}}$,
\begin{equation}
\FA=\frac{1}{|F_{\mathrm{eval}}|}\sum_{(x_i,y_i)\in F_{\mathrm{eval}}}
\mathbf{1}\{\Mhat(x_i)=y_i\}.
\end{equation}
All questions are multiple choice. The prompt lists the options, and $\Mhat(x_i)$ is the option letter with the highest next-token logit. Lower FA means the model gives the target answer less often. FA measures answer suppression; it does not show that the information is gone.

\paragraph{Retain Accuracy (RA)}
For retained examples $R_{\mathrm{eval}}$,
\begin{equation}
\RA=\frac{1}{|R_{\mathrm{eval}}|}\sum_{(x_i,y_i)\in R_{\mathrm{eval}}}
\mathbf{1}\{\Mhat(x_i)=y_i\},
\end{equation}
where higher RA means better preserved performance on retained data.

\paragraph{Membership Distinguishability ($\MIAadv$)}
Following confidence-based membership inference~\cite{shokri2017membership}, a logistic-regression attacker is trained on confidence features to separate forget examples from held-out test examples. A forget-set-only success rate is difficult to interpret when the attack performance crosses the chance threshold, because lower values do not always indicate lower distinguishability. We therefore compute balanced attack accuracy for each checkpoint,
\begin{equation}
\mathrm{BA}=\frac{\mathrm{TPR}+\mathrm{TNR}}{2},
\end{equation}
and measure its deviation from chance as
\begin{equation}
\MIAadv=2\left|\mathrm{BA}-0.5\right|.
\label{eq:miaadv}
\end{equation}
$\MIAadv=0$ means chance-level distinguishability, and larger values mean stronger separation. We compute it per checkpoint before averaging over seeds, so that deviations above and below chance do not cancel. $\MIAadv$ concerns this attack only; it is not a general privacy guarantee.

\paragraph{Activation Distance (AD)}
Outputs alone do not show how closely the unlearned model resembles $\Mstar$ internally. Let $a_M(x)$ denote the output of the second-to-last decoder layer of the language model of $M$ on the prompt for input $x$ (image and question tokens, no answer), averaged over all token positions. We define
\begin{equation}
\AD(\Mhat,\Mstar)=\frac{1}{|F_{\mathrm{eval}}|}\sum_{x\in F_{\mathrm{eval}}}
\left\|a_{\Mhat}(x)-a_{\Mstar}(x)\right\|_2.
\label{eq:ad}
\end{equation}
Lower AD means closer alignment with $\Mstar$ in activation space. AD does not show that the information is gone.

\paragraph{Jensen--Shannon Divergence (JS)}
We compare the predictive distributions of the unlearned and retain-only models with the Jensen--Shannon divergence~\cite{lin1991divergence}:
\begin{equation}
\JS(\Mhat,\Mstar)=\frac{1}{|F_{\mathrm{eval}}|}\sum_{x\in F_{\mathrm{eval}}}
\operatorname{JSD}\!\left(p_{\Mhat}(\cdot|x)\,\|\,p_{\Mstar}(\cdot|x)\right).
\label{eq:js}
\end{equation}
Lower JS means the two models give more similar predictions on the forget examples. The divergence is computed in double precision; examples whose half-precision probabilities are not finite are skipped and counted.

\subsection{Target: Distance to the Retrained Operating Point}
\label{sec:target}
Successful unlearning should leave the model where retraining without $F$ would have left it, not as far from the original model as possible. We therefore score every unlearned model by its distance to the retrained operating point,
\begin{equation}
D(\Mhat)=\left|\FA(\Mhat)-\FA(\Mstar)\right|+\left|\RA(\Mhat)-\RA(\Mstar)\right|,
\label{eq:target}
\end{equation}
where $\FA(\Mstar)$ and $\RA(\Mstar)$ are those of the seed-42 retrained model. Lower $D$ is better. A criterion that is read as lower-is-better should then correlate positively with $D$. Because $D$ is built from FA and RA, part of their association with $D$ is mechanical; AD, JS, and $\MIAadv$ do not enter $D$, and their associations are not.

Together, these criteria examine different parts of the evaluation problem. FA measures direct forgetting, RA measures retained utility, $\MIAadv$ measures distinguishability under the evaluated membership attack, while AD and JS measure proximity to the same retain-only reference in activation and predictive-distribution space. Supplementary Table~S8 summarizes both what each criterion measures and what cannot be concluded from that criterion alone.

\paragraph{Scalarization}
\label{sec:uqs}
Our target $D$ is itself a utility choice: it weights forget and retain deviations equally and ignores representations. We use it to assess criteria, not as a universal method-selection objective; any scalar ranking should declare its deployment utility and report its sensitivity (supplementary material).

\subsection{Validity Conditions}
\label{sec:validity}

The criteria above are only meaningful if the pipeline that produced them satisfies three conditions. We state each as a test.

\emph{C1, conditioning.} The original model must have memorized the forget set, the original and retain-only models must have similar retained accuracy, and the forget and retain sets must be free of duplicate leakage. Otherwise, a low FA after unlearning may reflect model damage or an invalid data split rather than forgetting. We require
\begin{equation}
\FA(M)-\FA(\Mstar)\geq \gamma,\qquad
|\RA(M)-\RA(\Mstar)|\leq 0.03,
\label{eq:gate}
\end{equation}
with $\gamma=0.15$, and require that no forget question has a duplicate in the retain set. We also report the forget-set gap on held-out variants (Section~\ref{sec:heldout}) to test whether memorization transfers beyond the exact training questions and answer-option order. The threshold is a design choice, not a statistical constant. With 100 forget questions, the standard error of the difference between two accuracies is approximately 0.04--0.05, so $\gamma=0.15$ represents a gap of roughly three standard errors. Section~\ref{sec:conditioning} reports how the decisions change with $\gamma$.

\emph{C2, measurement.} The scorer must separate correct from incorrect answers. A scorer that gives a wrong answer nearly the same score as the gold answer measures nothing.

\emph{C3, execution.} The training code must implement the intended procedure. We test this condition by reintroducing two known bugs, one at a time, and checking whether forget-specific memorization survives.
\section{Experiments}
\label{sec:protocol}

The experiments follow the order of the contributions. Sections~\ref{sec:conditioning}--\ref{sec:execution} test the three validity conditions. Section~\ref{sec:agreement} examines whether each criterion tracks the distance to the retrained model. Section~\ref{sec:dose} studies how this relationship changes across operating regimes. Section~\ref{sec:donothing} compares the unlearned models with the original model. Section~\ref{sec:backbone} tests whether the findings are consistent across different backbones. Sections~\ref{sec:heldout} and~\ref{sec:shuffle} examine whether the findings reflect memorized training questions or the underlying knowledge. We first describe the experimental setup.

\subsection{Experimental Setup}
\subsubsection{Models and adaptation}
The primary model is LLaVA-1.5-7B~\cite{liu2023llava,liu2024visual} adapted with LoRA~\cite{hu2022lora}. The second backbone is BakLLaVA~\cite{bakllava2023}, which pairs the same visual design with a Mistral-7B language model~\cite{jiang2023mistral}. The third is LLaVA-Interleave-Qwen-7B~\cite{li2024llavanext}, which keeps the LLaVA model class but uses a Qwen1.5-7B language model and a SigLIP vision encoder. All use LoRA rank $r=8$, $\alpha=16$, and adapters on the q/k/v/o projection modules of the language model (about 0.1\% of the parameters; 8.0M for BakLLaVA and 9.8M for Qwen). The loss is computed on answer tokens only; prompt and image tokens are masked. All training uses AdamW, FP16, and gradient checkpointing on one NVIDIA RTX 4090 (24 GB); unlearning runs for one epoch with batch size 2 and gradient clipping at 1.0.

\subsubsection{Benchmarks}
We use MMBench-EN~\cite{liu2024mmbench} and the image questions of ScienceQA~\cite{lu2022scienceqa}. MMBench uses circular evaluation, so its 4{,}377 labelled rows contain only 1{,}175 unique questions. We de-duplicate both benchmarks by a content key (question text, sorted options, and a $32\times32$ grayscale thumbnail of the image), shuffle with seed 42, and draw 1{,}000 training questions (100 forget, 900 retain). The held-out test set has 175 questions for MMBench and 200 for ScienceQA, the latter drawn from the validation split (2{,}097 rows, 2{,}087 unique).

The original model $M$ is LoRA-fine-tuned on $F\cup R$ for eight epochs at learning rate $2\times10^{-5}$; the retrained reference $\Mstar$ uses the same recipe on $R$ only. MLLMU-Bench~\cite{liu2025mllmu} was trained with the same recipe but failed the memorization gate (Section~\ref{sec:conditioning}) and is excluded from the main study.

\subsubsection{Compared unlearning methods}
Six methods are applied to $M$ for one epoch. GA, RL, and FT-Retain follow the standard baselines of classification unlearning~\cite{jia2023sparsity,fan2024salun}. Gradient Ascent (GA) maximizes the loss on forget examples and nothing else. Gradient Difference (GradDiff) adds a retain term, $-\mathcal{L}_F+0.5\,\mathcal{L}_R$, as in TOFU~\cite{maini2024tofu}. Random Labels (RL) fine-tunes on forget questions whose answer is replaced by a wrong option of the same question. FT-Retain continues optimization only on retained data. SalUn~\cite{fan2024salun} is our LoRA adaptation: a saliency mask keeps the half of the LoRA weights with the largest forget-loss gradient magnitude, and random-label forgetting plus a retain loss updates only those weights. NPO~\cite{zhang2024npo} uses $\beta=0.1$, the original model as reference, and a retain term.

\subsubsection{Design}
Each method is run at three learning rates, $10^{-4}$, $3\times10^{-5}$, and $10^{-5}$, giving 18 unlearned models per setting and seed. LLaVA and BakLLaVA use unlearning seeds $\{42,123,5508\}$ (54 models per benchmark); Qwen uses seed 42. For each backbone and benchmark, one original model and one retrained model are trained with seed 42; the unlearning seeds change only the unlearning run. Qwen on ScienceQA misses the memorization gate (Section~\ref{sec:conditioning}) and is not unlearned, so the main study contains 234 unlearned models in five backbone$\times$benchmark settings. Section~\ref{sec:shuffle} adds 36 models from LLaVA trained with shuffled option orders.

\subsubsection{Statistical analysis}
Within each setting we compute the Spearman correlation between each criterion and $D$ (Eq.~\eqref{eq:target}). The three seeds of one method and learning rate are not independent, so 95\% intervals come from a cluster bootstrap (4{,}000 resamples) over the 18 method$\times$learning-rate configurations; with one seed the bootstrap reduces to resampling models. The per-setting estimates are the primary evidence. The five settings are not independent samples: the backbones use the same forget, retain, and test questions, and the methods and learning rates are shared. We therefore use the pooled estimate as a summary of the five settings and not as an estimate for a population of settings. Its $p$-values and intervals assume independent settings and are exploratory. We check it three ways. First, we count in how many settings the sign agrees with the pooled sign. Second, we repeat the pooling with every setting weighted as if it had only 18 observations. Third, we leave out each setting in turn. The pooling itself is a DerSimonian--Laird random-effects meta-analysis on Fisher-$z$ values~\cite{dersimonian1986meta}. Method rankings are compared with Kendall $\tauB$.

\subsubsection{Code availability}
Code, result files, and scripts for every table and figure are available at \url{https://github.com/abdullahak07/ValidityGate}.

\subsection{Benchmark Conditioning}
\label{sec:conditioning}

Table~\ref{tab:gate} reports the memorization gate. LLaVA and BakLLaVA pass on both benchmarks and Qwen on MMBench, with gaps of 0.18--0.21 (3.8--5.0 standard errors) while retain accuracy changes by at most 0.02. Qwen on ScienceQA misses the threshold with a gap of 0.14 and is not unlearned; we kept the rule fixed rather than adjust it after seeing the result. MLLMU-Bench does not: the original model answers 0.189 of the forget questions correctly and the retrained model 0.062, a gap of 0.127.

The exclusion of MLLMU-Bench depends on $\gamma$, and we report it that way. Its gap is 2.8 standard errors, so the original model has memorized something; at $\gamma\leq0.12$ it would pass, while our five main settings pass at every $\gamma\leq0.18$. The stronger reason for excluding it is the floor. The original model answers only 18.9\% of the forget questions, so the whole range available to an unlearning method, from the original model to the retrained one, is 0.127. Differences between methods inside that range are of the same size as the sampling error of FA on 100 questions, and FA cannot rank them. This is the situation ReMem warns about~\cite{kwon2026remember}: the targets were barely learned in the first place.

\begin{table}[!htbp]
\centering
\small
\caption{Memorization gate (Eq.~\eqref{eq:gate}). FA and RA of the original model $M$ and the retrained reference $\Mstar$; all models use seed 42. A setting passes when the forget gap is at least $\gamma=0.15$ and RA changes by at most 0.03; $z$ is the gap divided by its binomial standard error ($n=100$). $^\dagger$Fails at $\gamma=0.15$; passes at $\gamma\leq0.12$. $^\ddagger$Fails at $\gamma=0.15$; not unlearned. LLaVA-shuf.: option-shuffled training (Section~\ref{sec:shuffle}).}
\label{tab:gate}
\setlength{\tabcolsep}{3pt}
\begin{tabular}{llccccc}
\toprule
\textbf{Backbone} & \textbf{Benchmark} & \multicolumn{2}{c}{$M$} & \multicolumn{2}{c}{$\Mstar$} & \textbf{Gap ($z$)}\\
 & & FA & RA & FA & RA & \\
\midrule
LLaVA & MMBench & 0.99 & 0.95 & 0.81 & 0.96 & 0.18 (4.4)\\
LLaVA & ScienceQA & 0.95 & 0.92 & 0.77 & 0.90 & 0.18 (3.8)\\
BakLLaVA & MMBench & 1.00 & 1.00 & 0.80 & 1.00 & 0.20 (5.0)\\
BakLLaVA & ScienceQA & 0.98 & 0.95 & 0.77 & 0.94 & 0.21 (4.7)\\
Qwen & MMBench & 1.00 & 1.00 & 0.82 & 1.00 & 0.18 (4.7)\\
Qwen & ScienceQA & 0.99 & 0.99 & 0.85 & 0.99 & 0.14 (3.8)$^\ddagger$\\
LLaVA-shuf. & MMBench & 1.00 & 0.98 & 0.85 & 0.99 & 0.15 (4.2)\\
LLaVA-shuf. & ScienceQA & 0.98 & 0.97 & 0.82 & 0.98 & 0.16 (3.9)\\
\midrule
LLaVA & MLLMU & 0.189 & -- & 0.062 & -- & 0.127 (2.8)$^\dagger$\\
\bottomrule
\end{tabular}
\end{table}

Duplicates are a second conditioning failure. Because each MMBench question appears in about 3.7 option-permuted copies, a random 1{,}000-row split without de-duplication places a duplicate of about half of the forget questions in the retain set. The model is then trained on the questions it is asked to forget, and FA after unlearning measures little. 

\subsection{Measurement Validity}
\label{sec:measurement}

Table~\ref{tab:scorer} summarizes the scorer audit. When the prompt lists the options, three scoring rules (letter logits, a strict rule that accepts a generated answer only if it starts with the gold letter or equals the gold option text, and a lenient rule that accepts it if the gold letter appears anywhere) rank the 20 seed-42 models of each benchmark almost identically ($\tauB$ between 0.89 and 1.0). The lenient rule fails when the options are removed from the prompt, as some pipelines do. The untrained base model on MMBench then scores 0.87 against the gold letter and 0.84 against a wrong letter of the same question, while a match on the gold option text gives 0.15. A scorer that cannot tell the gold letter from a wrong one carries no information about forgetting.

The failure changes conclusions, not only numbers. We applied a no-options lenient scorer to the same correctly trained seed-42 models. The memorization gap between the original and retrained models falls from 0.18 to 0.04 on MMBench and from 0.18 to $-0.01$ on ScienceQA, so both benchmarks would fail the gate and be read as never memorized. Across the 12 unlearned models at $10^{-4}$ and $3\times10^{-5}$, the lenient score of the gold letter exceeds that of a wrong letter by 0.008 on average on MMBench (0.045 on ScienceQA), and the spread of FA shrinks from 0.18--0.97 to 0.44--0.87. Even with the options in the prompt, the lenient rule overstates FA when unlearning degrades the output format: NPO at $10^{-4}$ on MMBench scores 0.90 instead of 0.67, because its answers list several letters.

Free-text identity questions show a related problem. On MLLMU-Bench, token F1 separates the original and retrained models by 0.05, against 0.14 when an answer is scored by whether it names the right person. On LLaVA-1.5-7B adapters trained to memorize the MLLMU forget profiles (37 forget biographies with a usable name plus 60 retain biographies, against a reference trained on the 60 retain biographies alone; 16 epochs at $2\times10^{-4}$), identity scoring gives 0.622 vs.\ 0.000, and token F1 gives 0.452 vs.\ 0.276. Token F1 rewards the shared wording of biographies and hides most of the memorization signal.

\begin{table}[!htbp]
\centering
\small
\caption{Scorer audit. ``Placebo'' applies the lenient rule to a wrong letter of the same question; an informative scorer must score it much lower than the gold letter.}
\label{tab:scorer}
\setlength{\tabcolsep}{4pt}
\begin{tabular}{L{3.6cm}cc}
\toprule
\textbf{Check} & \textbf{Value} & \textbf{Reading}\\
\midrule
Rank agreement of three scorers, options in prompt ($\tauB$) & 0.89--1.0 & consistent\\
Lenient, no options: gold / placebo (base, MMBench) & 0.87 / 0.84 & uninformative\\
Option-text match, no options (base, MMBench) & 0.15 & informative\\
Gate gap, letter / lenient, same models (MMBench) & 0.18 / 0.04 & gate hidden\\
Gate gap, letter / lenient, same models (ScienceQA) & 0.18 / $-0.01$ & gate hidden\\
NPO $10^{-4}$ FA, letter / lenient with options & 0.67 / 0.90 & overstated\\
MLLMU gap, token F1 / identity (full models) & 0.05 / 0.14 & F1 hides gap\\
MLLMU memorized adapters, identity ($M$ / $\Mstar$) & 0.622 / 0.000 & clear\\
MLLMU memorized adapters, token F1 ($M$ / $\Mstar$) & 0.452 / 0.276 & compressed\\
\bottomrule
\end{tabular}
\end{table}

\subsection{Execution Validity}
\label{sec:execution}

For bug injection, a fresh LoRA adapter is trained on the 100 ScienceQA forget questions and 60 retain questions, and forget and held-out test accuracy are compared with the untrained model. The forget-specific gain is the forget gain minus any positive test gain. Each condition switches one bug back on (Table~\ref{tab:bugs}). The correct pipeline gives a gain of $+0.31$. Training for one epoch gives $+0.02$, and computing the loss over prompt and image tokens gives $-0.04$: both remove the memorization that every later unlearning score depends on, and both runs finish without an error. Two other suspected bugs had no effect. A reentrant gradient-checkpointing configuration that we expected to block gradients to the adapter gave $+0.32$ in the current library versions, and a misconfigured cosine schedule gave $+0.32$. We report these as non-findings. The memorization gate in Eq.~\eqref{eq:gate} catches both harmful bugs.

\begin{table}[!htbp]
\centering
\small
\caption{Bug injection on ScienceQA (LLaVA-1.5-7B). Forget-specific gain $=$ forget-accuracy gain $-\max(0,$ test-accuracy gain$)$ relative to the untrained model.}
\label{tab:bugs}
\begin{tabular}{lc}
\toprule
\textbf{Pipeline} & \textbf{Forget-specific gain}\\
\midrule
Correct pipeline & $+0.31$\\
One training epoch & $+0.02$\\
Loss over the whole sequence & $-0.04$\\
Reentrant checkpointing & $+0.32$\\
Cosine schedule stepped per batch & $+0.32$\\
\bottomrule
\end{tabular}
\end{table}

\subsection{Do the Criteria Track the Retrained Operating Point?}
\label{sec:agreement}

Table~\ref{tab:corr} (forest plot in supplementary Fig.~S1) reports the association of each criterion with $D$. Forget accuracy has a negative association in four of the five settings and none on BakLLaVA ScienceQA ($-0.02$). The exploratory pooled estimate is $\rho=-0.51$ (95\% CI $[-0.70,-0.25]$, $p=0.0004$); the conservative pooling in Table~\ref{tab:corr} ($-0.54$) and leave-one-setting-out estimates (between $-0.59$ and $-0.42$, $p\leq0.007$) agree in sign. The settings differ ($I^2=0.81$): the effect is strongest on Qwen ($-0.83$). Read as lower-is-better, FA would therefore favor the models farthest from the retrained operating point. The per-method results show where this comes from (supplementary Tables S1--S7): at $10^{-4}$ most methods push FA far below the retrained value $\FA(\Mstar)$ and pull RA down with it, and the negative association disappears once these collapsed models are removed (Section~\ref{sec:dose}).

\begin{table*}[!t]
\centering
\small
\caption{Spearman $\rho$ between each criterion and the distance $D$ to the retrained operating point, with 95\% bootstrap intervals (cluster bootstrap over method$\times$learning-rate configurations), and the random-effects pooled estimate over the five main settings. $^{\ast}$Pooled rows are exploratory because the settings share questions and methods; the conservative pooling weights each setting as $n=18$. $I^2$ is between-setting heterogeneity. The last two rows (option-shuffled training, Section~\ref{sec:shuffle}) are not pooled. For a lower-is-better criterion, a positive $\rho$ means it tracks $D$.}
\label{tab:corr}
\footnotesize
\setlength{\tabcolsep}{2.4pt}
\begin{tabular}{lccccc}
\toprule
\textbf{Setting} ($n$) & \textbf{FA} & \textbf{1$-$RA} & \textbf{$\MIAadv$} & \textbf{AD} & \textbf{JS}\\
\midrule
LLaVA MMBench (54) & $-0.59$ [$-0.90$, $-0.00$] & $+0.67$ [$+0.18$, $+0.92$] & $-0.14$ [$-0.52$, $+0.25$] & $+0.59$ [$+0.23$, $+0.81$] & $+0.43$ [$-0.05$, $+0.73$]\\
LLaVA ScienceQA (54) & $-0.51$ [$-0.87$, $+0.11$] & $+0.62$ [$+0.11$, $+0.89$] & $-0.23$ [$-0.44$, $+0.00$] & $+0.54$ [$+0.14$, $+0.82$] & $+0.58$ [$+0.23$, $+0.79$]\\
BakLLaVA MMBench (54) & $-0.50$ [$-0.82$, $+0.07$] & $+0.68$ [$+0.29$, $+0.89$] & $+0.02$ [$-0.30$, $+0.35$] & $+0.66$ [$+0.28$, $+0.86$] & $+0.33$ [$-0.06$, $+0.63$]\\
BakLLaVA ScienceQA (54) & $-0.02$ [$-0.55$, $+0.65$] & $+0.23$ [$-0.29$, $+0.66$] & $-0.41$ [$-0.64$, $-0.12$] & $+0.21$ [$-0.27$, $+0.62$] & $+0.07$ [$-0.46$, $+0.54$]\\
Qwen MMBench (18) & $-0.83$ [$-0.97$, $-0.46$] & $+0.85$ [$+0.52$, $+0.96$] & $+0.30$ [$-0.30$, $+0.71$] & $+0.85$ [$+0.52$, $+0.97$] & $+0.32$ [$-0.25$, $+0.90$]\\
\midrule
Pooled$^{\ast}$ & $-0.51$ [$-0.70$, $-0.25$] & $+0.63$ [$+0.42$, $+0.77$] & $-0.14$ [$-0.33$, $+0.06$] & $+0.58$ [$+0.38$, $+0.74$] & $+0.36$ [$+0.17$, $+0.53$]\\
$p$ / $I^2$ & 0.0004 / 0.81 & $<10^{-4}$ / 0.76 & 0.177 / 0.55 & $<10^{-4}$ / 0.75 & 0.0004 / 0.57\\
Conservative$^{\ast}$ & $-0.54$ [$-0.75$, $-0.23$] & $+0.65$ [$+0.42$, $+0.79$] & $-0.10$ [$-0.33$, $+0.15$] & $+0.61$ [$+0.37$, $+0.77$] & $+0.36$ [$+0.14$, $+0.54$]\\
\midrule
LLaVA-shuf.\ MMBench (18) & $-0.46$ [$-0.85$, $+0.19$] & $+0.58$ [$+0.04$, $+0.90$] & $+0.35$ [$-0.24$, $+0.83$] & $+0.43$ [$-0.06$, $+0.75$] & $+0.22$ [$-0.31$, $+0.64$]\\
LLaVA-shuf.\ ScienceQA (18) & $-0.56$ [$-0.88$, $+0.09$] & $+0.66$ [$+0.11$, $+0.93$] & $+0.09$ [$-0.43$, $+0.63$] & $+0.62$ [$+0.20$, $+0.87$] & $+0.58$ [$+0.12$, $+0.81$]\\
\bottomrule
\end{tabular}
\end{table*}

Membership distinguishability does not track the target. The pooled association is $\rho=-0.14$ ($[-0.33,+0.06]$, $p=0.18$), its sign changes across settings (from $-0.41$ on BakLLaVA ScienceQA to $+0.30$ on Qwen), and only one per-setting interval excludes zero. Reference rows show why the statistic is hard to read at this scale: the retrained model, which never saw the forget set, is more distinguishable than the original model on BakLLaVA MMBench (0.19 vs.\ 0.01) and on Qwen MMBench (0.13 vs.\ 0.04). A rule that prefers lower $\MIAadv$ would call the original model more forgotten than the reference it should match. This result concerns the confidence-based attack evaluated here, not membership privacy in general; per-example attacks are known to reveal leakage that population attacks miss~\cite{hayes2025inexact}.

The reference-based criteria behave as expected. AD has the most consistent association ($\rho=+0.58$, between $+0.52$ and $+0.64$ when any one setting is left out), and JS follows ($+0.36$, between $+0.29$ and $+0.44$). Neither enters the definition of $D$, so these associations are not mechanical. Both require the retrained model, which a deployed system does not have, so they can audit an unlearning method but cannot replace retraining. Retain loss $1-\RA$ has the largest pooled association ($+0.63$), in part because RA enters $D$ directly. Within the non-collapsed models these associations disappear (Section~\ref{sec:dose}), so they mainly separate collapsed from healthy models.

Because FA itself enters $D$, its negative association could partly reflect how $D$ is built. We therefore repeat the test against two targets that do not contain FA: the activation distance AD and the output divergence JS between each unlearned model and the retrained model on the forget set (Table~\ref{tab:indep}). The direction holds and becomes stronger. Lower FA goes with larger AD in every setting ($\rho$ from $-0.59$ to $-0.95$, all intervals excluding zero) and with larger JS in every setting ($-0.44$ to $-0.80$; the interval includes zero only for Qwen, whose collapsed models output nearly constant distributions). This includes BakLLaVA ScienceQA, where $D$ showed no association ($-0.76$ for both targets). The weighting inside $D$ matters for the $D$-based result: when $D$ uses forget deviation only, the association weakens on LLaVA and BakLLaVA and its intervals include zero, so part of the $D$-based effect comes from retain accuracy falling together with FA at large learning rates. The conclusion that lower FA does not mean closer to the retrained model rests on the independent targets, and only on these targets does it survive the removal of collapsed models (Section~\ref{sec:dose}).

\begin{table*}[!t]
\centering
\footnotesize
\caption{Forget accuracy against targets that do not contain FA (JS and AD to the retrained model on the forget set), and against $D$ with forget weight $w$, $D_w=w|\FA-\FA^*|+(1-w)|\RA-\RA^*|$ ($w=0.5$ is Eq.~\eqref{eq:target} up to a factor). Spearman $\rho$ with 95\% cluster-bootstrap intervals. A negative value means lower FA goes with a larger distance from the retrained model.}
\label{tab:indep}
\setlength{\tabcolsep}{6pt}
\begin{tabular}{lccc}
\toprule
\textbf{Setting} & \textbf{FA vs.\ JS} & \textbf{FA vs.\ AD} & \textbf{FA vs.\ $D_{w=1}$}\\
\midrule
LLaVA MMB & $-0.78$ [$-0.92$, $-0.54$] & $-0.74$ [$-0.92$, $-0.40$] & $-0.38$ [$-0.80$, $+0.25$]\\
LLaVA SQA & $-0.78$ [$-0.88$, $-0.53$] & $-0.71$ [$-0.88$, $-0.41$] & $-0.20$ [$-0.69$, $+0.48$]\\
BakLLaVA MMB & $-0.64$ [$-0.84$, $-0.33$] & $-0.84$ [$-0.93$, $-0.57$] & $-0.32$ [$-0.76$, $+0.38$]\\
BakLLaVA SQA & $-0.76$ [$-0.90$, $-0.46$] & $-0.76$ [$-0.90$, $-0.40$] & $+0.07$ [$-0.48$, $+0.71$]\\
Qwen MMB & $-0.44$ [$-0.96$, $+0.13$] & $-0.95$ [$-0.99$, $-0.80$] & $-0.83$ [$-0.98$, $-0.43$]\\
LLaVA-shuf.\ MMB & $-0.55$ [$-0.86$, $-0.06$] & $-0.82$ [$-0.93$, $-0.57$] & $-0.27$ [$-0.77$, $+0.39$]\\
LLaVA-shuf.\ SQA & $-0.80$ [$-0.95$, $-0.47$] & $-0.59$ [$-0.84$, $-0.19$] & $-0.45$ [$-0.84$, $+0.33$]\\
\bottomrule
\end{tabular}
\end{table*}

\subsection{Operating Regime and Method Selection}
\label{sec:dose}

\begin{figure*}[!t]
\centering
\includegraphics[width=\textwidth]{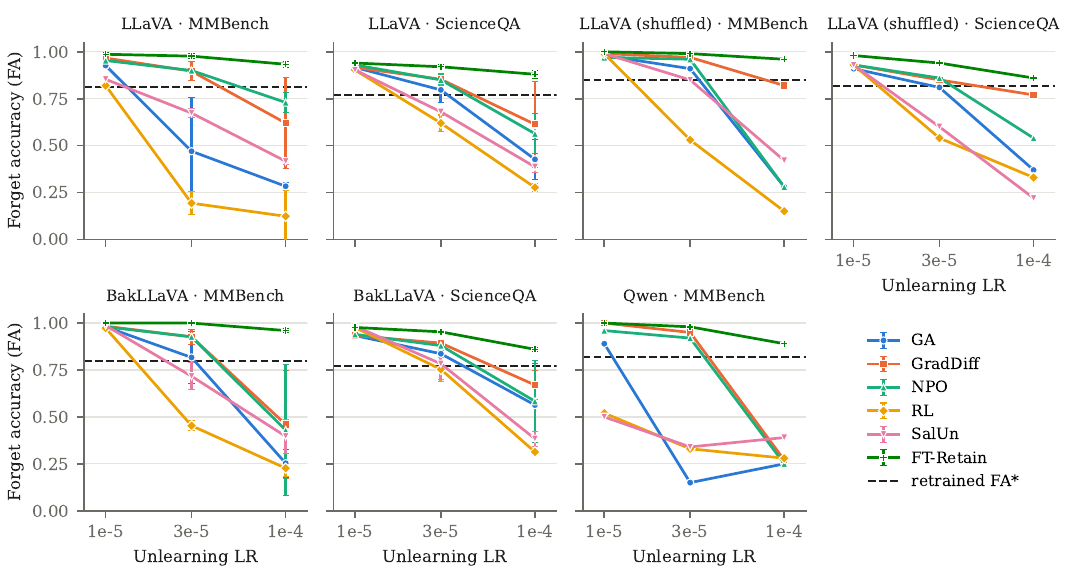}
\caption{Forget accuracy against the unlearning learning rate for each method, backbone, and benchmark. The dashed line marks the retrained forget accuracy $\FA(\Mstar)$. LLaVA and BakLLaVA points are means over three seeds with $\pm1$ SD; Qwen and the option-shuffled LLaVA models have one seed. Large learning rates overshoot $\FA(\Mstar)$, which is where lower FA stops meaning closer to the retrained operating point.}
\label{fig:dose}
\end{figure*}

The learning rate acts as a dose (Fig.~\ref{fig:dose}). At $10^{-5}$ most methods barely move from the original model and FA stays above $\FA(\Mstar)$. At $10^{-4}$ GA, RL, SalUn, NPO, and GradDiff overshoot, and FA falls far below the retrained value while RA follows it down. The models closest to the retrained operating point sit at intermediate doses. Qwen is more fragile: RL and SalUn already collapse at $10^{-5}$ (FA and RA near 0.5), so its safe dose range is narrower than that of the other backbones.

The direction of FA therefore depends on the regime. Among the six method means for LLaVA, the FA ranking and the $D$ ranking agree at $10^{-5}$ ($\tauB=+0.87$ on MMBench, $+0.96$ on ScienceQA). They are reversed at $10^{-4}$ ($\tauB=-1.00$ and $-0.87$). In both benchmarks RL has the lowest FA at every learning rate, yet at $10^{-4}$ it is the method farthest from the retrained operating point ($D=1.49$ on MMBench). The method with the lowest $D$ changes with the dose: FT-Retain at $10^{-4}$, GradDiff at $3\times10^{-5}$, and SalUn (MMBench) or RL and SalUn (tied, ScienceQA) at $10^{-5}$. The $D$ rankings at $10^{-4}$ and $10^{-5}$ are anti-correlated ($\tauB=-0.60$ and $-0.55$). A comparison of methods at a single learning rate thus says as much about the chosen dose as about the methods. GEMMU independently shows that multimodal-unlearning rankings depend on method-specific hyperparameters~\cite{chew2026gemmu}; our complementary result identifies the direction of the failure relative to a retrained counterfactual: aggressive updates overshoot $\FA(\Mstar)$ and enter a collapsed regime. The supplementary material gives FA, RA, $\MIAadv$, AD, JS, and $D$ for every method and learning rate.

\paragraph{Collapse-Stratified Analysis}
Pooling healthy and collapsed models could create the negative association by itself, so we separate them by retain accuracy (supplementary Table~S10). Of the 234 models, 64 fall below $\RA=0.70$. All of them were trained at $3\times10^{-5}$ or $10^{-4}$, except Qwen's RL and SalUn models, and their mean FA is 0.30--0.42. Among the collapsed models, FA and $D$ are almost perfectly anti-correlated ($\rho$ from $-0.88$ to $-0.99$). Among the 170 healthy models the sign reverses: FA is positively associated with $D$ (pooled $\rho=+0.47$, $[+0.17,+0.69]$, $p=0.003$, $I^2=0.76$; conservative pooling $+0.50$, $[+0.15,+0.74]$); the per-setting intervals exclude zero only on BakLLaVA ScienceQA and Qwen. The sign does not depend on the threshold: from $\RA\geq0.60$ to $\RA\geq0.80$ the pooled healthy-model estimate ranges from $+0.37$ to $+0.77$, and the do-nothing comparison of Section~\ref{sec:donothing} moves by a few points at most (75--84\% closer on $D$, 93--94\% farther on AD). The negative pooled association of Table~\ref{tab:corr} is therefore a property of the mix of regimes, not of FA within a regime.

With the stricter filter $\RA\geq\RA(\Mstar)-0.10$ the healthy-model association rises to $+0.89$, but this filter removes most of the RA term from $D$, so that value is close to mechanical. Most healthy models stay above $\FA(\Mstar)$ (6--7 per LLaVA or BakLLaVA setting fall below it, none on Qwen), so lowering FA brings them toward the retrained forget accuracy. FA can be read as lower-is-better for the operating point only once collapse is excluded, which itself requires $\FA(\Mstar)$ or the learning-rate curve.

The independent targets behave differently from $D$. Within the healthy models AD and JS no longer track $D$ (pooled $-0.07$ and $-0.19$), and $\MIAadv$ still does not ($-0.23$, $[-0.44,+0.01]$). Yet among healthy models, lower FA still goes with larger AD (pooled $-0.57$, $[-0.66,-0.45]$, $I^2=0$) and larger JS ($-0.66$, $[-0.74,-0.56]$, $I^2=0$), with the same sign in every setting and intervals excluding zero on all four LLaVA and BakLLaVA settings. Healthy models that lower FA thus move toward the retrained forget accuracy while moving away from the retrained model's activations and predictions on the forget set: matching the aggregate accuracy is not the same as matching the model. Because every method starts from the original model, part of this association may reflect the size of the edit rather than its direction. Holding the learning rate fixed supports this caution: within $3\times10^{-5}$ the association remains ($-0.51$ for AD, $-0.81$ for JS), but within $10^{-5}$ and $10^{-4}$ it is not significant. The do-nothing comparison below does not rely on this correlation.

\subsection{Closer to Retraining than Doing Nothing?}
\label{sec:donothing}
Representation-level disagreement with output-level unlearning has been observed in image classification~\cite{kim2026truly,seo2025dimensional}; we ask whether it appears in multimodal LoRA unlearning when each unlearned model is compared with its own starting point. Every unlearning method starts from the original model $M$, which already lies at some distance from $\Mstar$. An unlearned model that ends farther from $\Mstar$ than $M$ was has done worse than running no unlearning at all. Fig.~\ref{fig:donothing} compares each healthy model ($\RA\geq0.70$) with $M$ on $D$, AD, and JS (counts in supplementary Table~S11). On $D$, unlearning mostly helps: 132 of the 170 models in the five main settings are closer to the retrained operating point than $M$ (78\%; between 70\% and 83\% per setting). On AD the picture reverses. Only 11 models are closer to $\Mstar$ than $M$, none on BakLLaVA, and none by more than 2.2\%; the median model is 13--31\% farther, depending on the setting. The option-shuffled models behave the same way (1 of 26 closer). JS is mixed: 65 of 170 models are closer, mostly on ScienceQA (49 of 85) rather than MMBench (16 of 85). The seeds and configurations are not independent, so we report these as counts rather than tests. The counts show that the methods do reach the retrained forget and retain accuracies, but they do not reach them by moving toward the retrained model in activation space. Whatever direction separates $M$ from $\Mstar$, gradient-based unlearning on the forget set does not find it. Combined with the recovery result of Section~\ref{sec:shuffle}, this is the strongest evidence in our data that matching the retrained scores is not the same as matching the retrained model.

\begin{figure*}[!t]
\centering
\includegraphics[width=\textwidth]{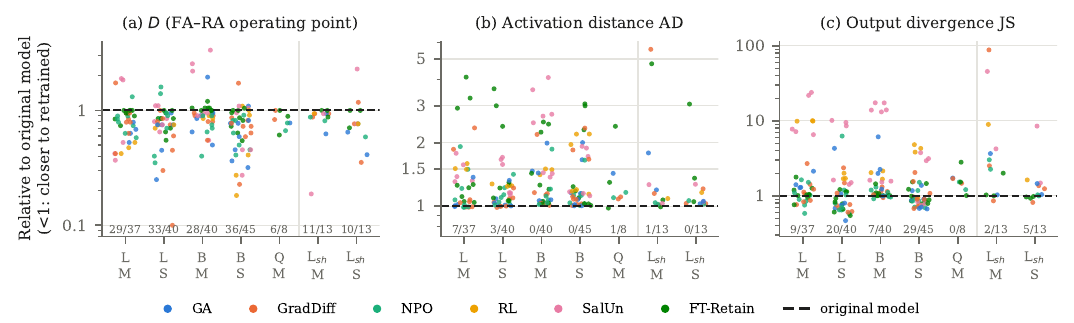}
\caption{Each healthy unlearned model ($\RA\geq0.70$) relative to the original model $M$, which receives no unlearning (dashed line at 1). Values below 1 mean the unlearned model is closer to the retrained model $\Mstar$ than $M$ is. (a) Distance $D$ to the retrained FA--RA operating point. (b) Activation distance AD. (c) Output divergence JS. Settings: L = LLaVA, B = BakLLaVA, Q = Qwen, L$_{sh}$ = LLaVA with option-shuffled training; M = MMBench, S = ScienceQA. Numbers give the models closer than $M$ out of all healthy models. Log scale; points are jittered horizontally.}
\label{fig:donothing}
\end{figure*}

\subsection{Second and Third Backbones}
\label{sec:backbone}

On BakLLaVA, now with three seeds, the MMBench results reproduce the LLaVA pattern: FA $\rho=-0.50$, $1-\RA$ $+0.68$, AD $+0.66$. On BakLLaVA ScienceQA, no criterion shows a clearly established association with $D$ except $\MIAadv$ ($-0.41$, $[-0.64,-0.12]$). The independent targets are clear in that setting, however (Table~\ref{tab:indep}), so the weak $D$-based result reflects how $D$ combines FA and RA there, not an absence of the effect. On the third backbone, Qwen on MMBench, the association is the strongest of all settings: FA $\rho=-0.83$, $1-\RA$ and AD $+0.85$. JS is weaker on BakLLaVA ($+0.33$, $+0.07$) and Qwen ($+0.32$) than on LLaVA ($+0.43$, $+0.58$); in BakLLaVA most JS values lie between 0.011 and 0.03, which leaves little range for ranking.

\subsection{Held-Out Variants}
\label{sec:heldout}

Forgetting and retention so far are measured on the questions used for unlearning, which GEMMU identifies as a pitfall~\cite{chew2026gemmu}. We therefore re-evaluated the models (evaluation only) on held-out variants: each forget question with its options rotated by one and by two positions, so the letters change and the contents do not, and with its options reversed; and the first 100 retain questions rotated by one. We also recorded, on the unseen test questions, how often each unlearned model gives the same answer as the retrained model.

\begin{table}[!htbp]
\centering
\footnotesize
\caption{Held-out evaluation. Gap: FA of $M$ minus FA of $\Mstar$ on the original and on option-rotated forget questions. $\rho$: Spearman correlation over the unlearned models between FA and (i) $D$ on rotated forget and retain questions, (ii) answer agreement with $\Mstar$ on the test set. Recovery: share of forget questions answered wrongly in the original order but correctly in every variant, unlearned models pooled vs.\ $\Mstar$. LLaVA: seed 42; BakLLaVA: three seeds. Shuffled models use the reversed order, the only one not trained (Section~\ref{sec:shuffle}).}
\label{tab:heldout}
\setlength{\tabcolsep}{1.8pt}
\begin{tabular}{lcccc}
\toprule
\textbf{Setting} & \textbf{Gap orig / rot.} & \textbf{$\rho$(FA, $D_{\text{ho}}$)} & \textbf{$\rho$(FA, agree)} & \textbf{Recovery}\\
\midrule
LLaVA MMB & 0.18 / 0.05 & $-0.84$ & $+0.95$ & 0.13 / 0.26\\
LLaVA SQA & 0.18 / 0.07 & $-0.60$ & $+0.93$ & 0.23 / 0.17\\
BakLLaVA MMB & 0.20 / 0.11 & $-0.57$ & $+0.91$ & 0.16 / 0.10\\
BakLLaVA SQA & 0.21 / 0.08 & $-0.67$ & $+0.93$ & 0.26 / 0.26\\
Qwen MMB & 0.18 / 0.15 & $-0.78$ & $+0.93$ & 0.04 / 0.00\\
Qwen SQA & 0.14 / 0.09 & -- & -- & --\\
\midrule
LLaVA-shuf.\ MMB & 0.15 / 0.14 & $-0.37$ & $+0.98$ & 0.09 / 0.07\\
LLaVA-shuf.\ SQA & 0.16 / 0.14 & $-0.54$ & $+0.94$ & 0.25 / 0.06\\
\bottomrule
\end{tabular}
\end{table}

Table~\ref{tab:heldout} gives three results for the main pipeline. First, models with lower FA agree less with the retrained model on test questions that no model was trained or unlearned on ($\rho$ between $+0.91$ and $+0.95$ in every setting). This target contains neither FA nor the forget set, and unlike $D$ it keeps its sign among healthy models ($+0.68$ to $+0.89$ in the five main settings), in line with the activation result of Section~\ref{sec:donothing}. Second, for LLaVA and BakLLaVA the memorization gap shrinks from 0.18--0.21 to 0.05--0.11 once the options are rotated: the original model's FA falls (for example from 0.99 to 0.88 for LLaVA on MMBench) while the retrained model's FA does not move. These models memorized mostly the specific training questions; Qwen kept more of the gap (0.15 on MMBench). Third, we find no sign of hidden knowledge in the main pipeline: unlearned models answer variants of forgotten questions correctly about as often as the retrained model does.

\subsection{Option-Shuffled Training}
\label{sec:shuffle}

The shrinking gap means that, under standard fine-tuning, LLaVA's forgetting results concern instances rather than knowledge. We therefore retrained the LLaVA original and retrained models on four option orders of every question (its cyclic rotations), with the number of epochs divided by four so that each question is still seen about eight times. The reversed order, which is never a rotation for three or more options, was never trained and serves as the held-out variant. Unlearning and FA/RA use the original questions; everything else is unchanged (36 unlearned models, seed 42).

Three results follow (Tables~\ref{tab:gate}, \ref{tab:corr}, \ref{tab:heldout}). First, the memorization is now about content: the gap is 0.15 and 0.16 on the training questions and 0.14 on the held-out reversed questions for both benchmarks, against 0.05--0.07 under standard training. Second, the regime pattern is unchanged: pooled over all models FA is negatively associated with $D$ ($-0.46$, $-0.56$) and with AD and JS (Table~\ref{tab:indep}), the sign turns positive once collapsed models are removed (supplementary Table~S10), almost no healthy model is closer to $\Mstar$ in activation space than the original model (Section~\ref{sec:donothing}), and models with lower FA agree less with the retrained model on test questions ($+0.98$, $+0.94$). Third, unlearning now leaves recoverable knowledge on ScienceQA. Of the forget questions an unlearned model answers wrongly in the original order, 25\% are answered correctly in every rotated and reversed order, against 6\% for the retrained model; at the two larger learning rates the rate lies between 7\% and 38\% and exceeds the retrained model's rate for every method. On MMBench there is no such effect (9\% vs.\ 7\%). When the knowledge itself is memorized, unlearning can suppress its original-format answer without removing it, which direct FA cannot detect.

\section{Discussion}
\label{sec:discussion}

\subsection{What the Evidence Supports}

The results support five conclusions. First, all three validity conditions can fail silently. Second, FA is informative only within the healthy regime. Aggressive updates can overshoot the retrained target, collapse RA, and reverse method rankings. Third, 132 of 170 healthy models move closer to the retrained FA--RA scores, but 159 move farther from the retrained model in activation space. Fourth, $\MIAadv$ does not track $D$. AD and JS track it more closely, but require a retrained reference and mainly separate healthy from collapsed models. Fifth, option-shuffled evaluation reveals knowledge that low FA does not detect. These results do not establish a universally correct metric or prove complete erasure.

\subsection{Why Lower Forget Accuracy Can Mislead}

The retrained model answers 77--85\% of the forget questions because some answers can be inferred from retained data. Reducing FA below this level therefore moves the unlearned model away from the retrained operating point. This occurs mainly at the largest learning rate, where RA also falls. FA should be interpreted relative to $\FA(\Mstar)$ and across several learning rates. Matching the retrained FA alone does not imply that the two models are equivalent.

\subsection{Why Pre-Unlearning Memorization Should Be Reported}

An unlearning score is not meaningful if the original model did not learn the target information. MLLMU-Bench fails our memorization gate, and both injected bugs remove memorization without producing an error. This agrees with ReMem~\cite{kwon2026remember}. Under standard fine-tuning, rotating the answer options reduces the forget-set gap from 0.18--0.21 to 0.05--0.11, consistent with GEMMU~\cite{chew2026gemmu}. Studies should therefore report the original-to-retrained forget-set gap on both training questions and held-out variants.

\subsection{Recommended Evaluation Protocol}

Before unlearning, verify memorization and remove duplicate leakage between the forget and retain sets. Audit the scorer using known correct, incorrect, and placebo answers. Compare each unlearned model with both the retrained reference and the original model as a do-nothing baseline. Evaluate several update strengths and interpret FA relative to the retrained value. Statistical tests should account for dependence among seeds and configurations. Any combined score should report its component metrics, weights, and sensitivity, and should not be presented as proof of erasure.

\subsection{Limitations}
\label{sec:limitations}

\emph{Statistical and reference scope.}
Each forget set contains 100 questions, so FA changes in steps of 0.01. Uncertainty is greatest for Qwen and the option-shuffled settings, which use one seed. The pooled estimates summarize five related settings and may not generalize beyond them. Because $D$ is calculated from FA and RA, their associations with $D$ are partly mechanical; this does not apply to AD, JS, or $\MIAadv$. The reference $\Mstar$ is one seed-42 retain-only adapted model, not retraining from initialization or a unique unlearning target.

\emph{Model and task scope.}
All backbones belong to the LLaVA family, all updates use LoRA, and all questions are multiple choice. The held-out evaluation changes the option order but retains the question and image. Paraphrased questions and new images would provide stronger tests of concept-level forgetting. The option-shuffled experiment also uses one backbone and one seed. Finally, the forget set contains fine-tuned knowledge, which may be easier to remove than knowledge acquired during pre-training~\cite{chew2026gemmu}.

\emph{Attack and metric scope.}
$\MIAadv$ represents one confidence-based attack; per-example attacks such as LiRA may behave differently~\cite{carlini2022membership,hayes2025inexact}. Option reordering is also a limited recovery test, and stronger probes may recover more information~\cite{lynch2024eight,zhang2025sua}. AD and JS increase with update magnitude, so their association with FA may reflect the size of the update as well as its direction. The healthy threshold $\RA=0.70$ is a design choice; the supplementary material also reports $\RA\geq\RA(\Mstar)-0.10$. \Fw{} is a technical evaluation method, not a test of legal compliance.

\section{Conclusion}
\label{sec:conclusion}

This paper proposed \Fw{}, which tests conditioning, measurement, and execution before interpreting multimodal unlearning results. Each condition can fail silently. After enforcing them, we evaluated 234 models across three LLaVA-family backbones. FA tracks the retrained operating point among healthy models but reverses after aggressive updates overshoot and collapse RA. Compared with doing nothing, 78\% of healthy models move closer to the retrained FA--RA scores, while 94\% move farther from the retrained model in activation space. After option-shuffled training, unlearned models answer 25\% of reordered forgotten ScienceQA questions, compared with 6\% for the retrained model. These results show that FA alone does not establish erasure. Future work should test additional model families, concept-level forgetting, stronger recovery attacks, and knowledge acquired during pre-training.

\section*{Acknowledgment}
The authors used AI-based writing assistance to edit the language of this manuscript. All technical content, experiments, analyses, and conclusions are the authors' own, and the authors take full responsibility for the text.

\begingroup
\scriptsize
\bibliographystyle{IEEEtran}
\bibliography{references}
\endgroup

\end{document}